\documentclass[11pt,a4paper]{article}
\usepackage[hyperref]{acl2019}
\usepackage{times}
\usepackage{latexsym}
\usepackage{graphicx}
\usepackage[export]{adjustbox}
\usepackage{url}

\usepackage{float}
\usepackage{array}
\usepackage{caption}
\usepackage{subcaption}
\usepackage{color}
\usepackage{tabularx}
\usepackage{multirow}
\usepackage{tikz}
\usepackage{tikz-qtree}
\usepackage{array}
\usepackage{float}
\usepackage{amsmath}
\usepackage{enumitem}
\restylefloat{table}

\usepackage[T1]{fontenc}
\usepackage{etoolbox}
 
\makeatletter
\patchcmd{\maketitle}{\@fnsymbol}{\@alph}{}{}  % Footnote numbers from symbols to small letters
\makeatother

\usepackage[utf8]{inputenc}
\usepackage{xcolor}
\usepackage{pgfplots}
    \usetikzlibrary{
        pgfplots.colorbrewer,
    }
    \pgfplotsset{
        cycle list/.define={my marks}{
            every mark/.append style={solid,fill=\pgfkeysvalueof{/pgfplots/mark list fill}},mark=*\\
            every mark/.append style={solid,fill=\pgfkeysvalueof{/pgfplots/mark list fill}},mark=square*\\
            every mark/.append style={solid,fill=\pgfkeysvalueof{/pgfplots/mark list fill}},mark=triangle*\\
            every mark/.append style={solid,fill=\pgfkeysvalueof{/pgfplots/mark list fill}},mark=diamond*\\
        },
    }
 \aclfinalcopy % Uncomment this line for the final submission
\usepackage{etoolbox}

\title{Natural Language Generation at Scale:\\ 
A Case Study for Open Domain Question Answering}

\author{
\textbf{Alessandra~Cervone$^{1}$\thanks{\textcolor{white}{s} Work done during internship at Amazon Alexa AI.}
},
\textbf{ Chandra~Khatri$^{2}$\thanks{\textcolor{white}{s} Work done while working at Amazon Alexa AI.}},
\textbf{ Rahul~Goel$^{3}$\thanks{\textcolor{white}{s} Work done while working at Amazon Alexa AI.},} \textbf{Behnam~Hedayatnia$^{4}$,} %\vspace{-.0em}
\\
\textbf{Anu~Venkatesh$^{4}$,} \textbf{Dilek~Hakkani-T{\"u}r$^{4}$,} \textbf{Raefer~Gabriel$^{4}$}
\\
$^1$Signals and Interactive Systems Lab, University of Trento, Italy\\
 $^2$Uber AI, $^3$Google, $^4$Amazon Alexa AI\\
  {\tt alessandra.cervone@unitn.it, chandrak@uber.com,}\\
  {\tt \
{\{behnam, anuvenk, hakkanit, raeferg\}}@amazon.com,} \\
{\tt goelrahul@google.com}\\
}

\date{}
\begin{document}
\maketitle
\begin{abstract}
Current approaches to Natural Language Generation (NLG) for dialog mainly focus
on domain-specific, task-oriented applications (e.g. restaurant booking) using
limited ontologies (up to 20 slot types), usually without considering the
previous conversation context. Furthermore, these approaches require large
amounts of data for each domain, and do not benefit from examples that may be
available for other domains.
% NOTE: I added usually, as Ondrej's work uses context.
% VP: ontologies, slot and types are confusing when presented all together (especially because they have not been introduced yet).
%This work explores the feasibility of statistical NLG for conversational applications with larger ontologies, which may be required by multi-domain dialog systems as well as open-domain question answering (QA) based on knowledge graphs. 
This work explores the feasibility of applying statistical NLG to scenarios
requiring larger ontologies, such as multi-domain dialog applications or
open-domain question answering (QA) based on knowledge graphs.
% VP: This work explores the feasibility of applying statistical NLG to scenarios requiring larger  ontologies, such as multi-domain dialog applications. 
We model NLG through an Encoder-Decoder framework using a large dataset of
interactions between real-world users and a conversational agent for open-domain
QA.
% VP: We focus on modeling NLG through an Encoder-Decoder framework for open-domain Question-Answering (QA). To Test our approach we use a large dataset of interactions between real-world users and a conversational agent.
%and using a Encoder-Decoder framework. 
First, we investigate the impact of increasing the number of slot types on the
generation quality and experiment with different partitions of the QA data with
progressively larger ontologies (up to 369 slot types).
% QUESTION: what is the number of frames or relations?
Second, we perform multi-task learning experiments between open-domain QA and task-oriented dialog, and benchmark our model on a popular NLG dataset.  
Moreover, we experiment with using the conversational context as an additional input to improve response generation quality.
% VP: Second, we benchmark our model on a popular NLG dataset and perform multi-task learning experiments between open-domain QA and task-oriented dialog. Last, we experiment with using conversational context as an additional input to improve response generation quality. 
Our experiments show the feasibility of learning statistical NLG models for
open-domain QA with larger ontologies.
\end{abstract}

%\vspace{0.3cm}
\section{Introduction}

\begin{table}
\scalebox{0.75}{
 \begin{tabular}{c c c c} 
 \hline
& \multicolumn{2}{c}{\textbf{Input}} & \textbf{Output}\\
\hline\hline
  & \textbf{context} & \textbf{MR} & \textbf{Text}  \\
 \hline\hline
 & & \textbf{inform} &  `fringale is a french  \\ 
 \textbf{Task} & - & \textit{name}: `fringale' & restaurant' \\  
 \textbf{Oriented}& & \textit{food}: `french' & `fringale serves \\
  & &  & french food' \\
 \hline
 & `when & \textbf{inform} & `1792' \\
  & was & \textit{timepoint}: '1792' & `kentucky formed \\ 
 \textbf{QA} & kentucky & \textit{objStr}: `kentucky' & in 1792' \\
 & founded' & \textit{claStr}: `state' & `kentucky founded  \\
 & & \textit{relStr}: 'founded' & in 1792' \\ 
 \hline
\end{tabular}
}
\caption{%\small 
Examples of input-output pairs from a task-oriented (Task) NLG
~(SFX~\cite{wen2015semantically}) and a Question-Answering (QA) dataset. In NLG
the input is typically a Meaning Representation~(MR) and the output is its
textual realization~(Text).  Each MR is composed of a Dialog Act (bold) and a
list of slot type (italic)-value pairs.  Compared to most NLG datasets, our QA
corpus also has the previous question (context) as input.  While in the
task-oriented setting we observe a one-to-one relation between slots in the
input and the ones realized in the text, the same is not true for QA.}
\label{tab:MRExample}
\end{table}

% This work investigates whether it is possible to extend State-of-the-art Natural Language Generation (NLG) approaches to applications which require substantially bigger ontologies. and a different application, i.e. open-domain Question Answering. 
%In this paper Natural Language Generation (NLG) is regarded as the task of generating text from non-linguistic input \citep{gatt2018survey}.
%\squeezeup
In dialog literature Natural Language Generation (NLG) is framed as the task of
generating natural language responses that faithfully convey the semantic
information given by a Meaning Representation~(MR). 
%First version
%A MR is typically a structure consisting of a Dialog Act~(DA) (e.g. ``inform'' in Table \ref{tab:MRExample}) and a list of associated slots organized as slot type-slot value pairs (e.g. \textit{food}:`french' in Table \ref{tab:MRExample}) representing the information which has to be conveyed in the generated text.
% Second version
A MR is typically a structure consisting of a Dialog Act~(DA) and a list of associated slots. While the DA \cite{stolcke2000dialogue,mezza2018iso} expresses the intent of the utterance to be generated (e.g. ``inform'' in Table
\ref{tab:MRExample}), the slots, organized as slot type-slot
value pairs (e.g. \textit{food}:`french' in Table \ref{tab:MRExample}), represent the information which has to be conveyed in the generated text.
%represent the concepts (typically entities) to be mentioned in the utterance.

% SOA: NLG 
% module in dialog literature (template-based approaches, then first neural approach, now SoA EncDec with attention). Limited ontologies (up to 20 slot types), given by focus on task-oriented domain. Recent E2E, better than previous datasets in many regards (style, size), but still restaurant domain with a small ontology. In most datasets the context is not considered (exception context).
So far statistical NLG for dialog has mainly been investigated in research for
task-oriented applications (e.g. restaurant reservation, bus information) in
narrow, controlled environments with limited ontologies, i.e. considering a
small set of DAs and slot types (respectively 12 and 8 in the popular San
Francisco restaurant dataset (SFX)~\cite{wen2015semantically}, 8 and 1 in the
recent E2E NLG challenge~\cite{novikova2017e2e}).
%~\cite{wen2015semantically,duvsek2016context}.
% DILEK: I introduced SFX earlier above. It may be good to replace all future references with SFX.
% VP: ...with limited ontology, i.e., considering a small range of  DAs and slots~\cite{wen2015semantically,duvsek2016context}.
% AC: followed VP version
% VP: are the slots referring to slot-types from before? or slot-type/slot-value pairs (i.e., entities)?
% AC: clarified mentioning specifically slot types
% Popular NLG datasets such as San Francisco restaurant~\cite{wen2015semantically} restrict their world to a set of 12 slot types and 8 DAs. Even the recent E2E NLG challenge~\cite{novikova2017e2e} presents an ontology of 8 slot types and 1 DA.
Furthermore, most datasets consider MRs in isolation~\cite{novikova2017e2e}
i.e., they lack conversational context, even though the previous utterances in
the dialog have been shown to improve the performance of task-oriented
NLG~\cite{duvsek2016context}.  These characteristics of current approaches to
NLG can be linked to the fact that a vast majority of dialog NLG research is
tested on a single domain where the dialog agent performs simple tasks such as
giving information about a restaurant, with few exceptions~\cite{wen2016multi}.

However, with the rise of conversational agents such as Amazon Alexa and Google
Assistant, there is an increasing interest in complex multi-domain tasks.  These
systems typically rely on hand-crafted NLG, but this approach cannot scale to
the complex ontologies which may be required in real-world applications (e.g.
booking a trip).
% VP: These systems typically rely on hand-crafted NLG but this approach cannot scale to the complex ontologies which may be required by real-world applications (e.g., booking a trip).
% AC: followed VP version

In this work we explore the applicability of current NLG models for
task-oriented dialog, based on a MR-to-text framework using Encoder-Decoder
architectures, to open-domain QA.  This allows us to investigate the performance
of current NLG research in an environment with (1) much larger numbers of slot
types, and (2) a different application compared to task-oriented dialog.
%, and (3) perform cross-domain experiments between open-domain QA and task-oriented dialog.
%We envision our approach as a first step towards an integrated statistical NLG module for a dialog system. Such a system should work with applications with a variety of ontology sizes and be flexible across multiple domains.
%Hence, we structure our NLG models for QA following the MR-to-text paradigm typical of task-oriented dialog using a neural Encoder-Decoder architecture.
% VP: Should Encoder Decoder have a - or / ?
% AC: fixed
We generate the QA datasets for our experiments using as source a large corpus
of open-domain QA pairs from interactions between real-world users and a
conversational agent.  For evaluation, we utilize both objective metrics and
human judgment.  We observe that NLG for open-domain QA poses its own challenges
compared to task-oriented dialog, since correct answers to the same question do
not necessarily convey all slot types in the MR (see Table~\ref{tab:MRExample}).

In particular, in our first set of experiments, we investigate the effect of
using increasingly larger ontologies with regards to slot types on the
performance of our NLG models for QA.  We find that, notwithstanding the larger
ontologies and the noisiness of our dataset, models' performance does not
degrade significantly in terms of naturalness of generated text and efficiency
in encoding the MR information (i.e. Slot Error Rate). Interestingly, we find it
improves for some of the human evaluation metrics.  We also observe that using
conversational context improves the quality of generated responses.
% to improve the quality of generated responses. % by measuring XX
%Rahul: Still not clear, we should say something like ^
%VP: not very meaningful, how does it improve the quality? based on what metrics?
% AC: should be clear now that sentences are rearranged
In our second set of experiments, we investigate whether jointly training NLG
models for task-oriented dialog and QA improves performances.  To this end, we
experiment with learning NLG models in a multi-task setting between our QA data
and SFX. Our experiments show that learning models in a multi-task setting lead
to better performances in terms of naturalness of the generated output for both
tasks.
%We assess our models using both objective metrics and human evaluation. We observe that NLG for open-domain QA poses its challenges compared to task-oriented dialog since correct answers to the same question do not necessarily convey all slot types in the MR (see Table~\ref{tab:MRExample}).

This work has several contributions:
\begin{enumerate} [ topsep=3pt,  parsep=2pt]
    \item We apply the MR-to-text framework (typical of NLG for task-oriented
      dialog) to a open-domain QA application.
    \item We explore the importance of adding the previous conversational
      context to improve the quality of the generated output.
    \item We investigate the possibility of learning NLG models using a
      MR-to-text approach with increasingly larger ontologies in terms of slot
      types.
    \item We experiment with multi-task learning for NLG between open-domain QA
      and task-oriented dialog.
    \item Finally we also propose new evaluation metrics (see Section
      \ref{sec:eval}) to capture the variability of output in open-domain QA
      compared to NLG for task-oriented dialog.
\end{enumerate}

% TODO: These 2 paras are not clearly written, the flow is not good. 
% Firstly, to study the scaling properties of current approaches, 
% we experiment with learning NLG models using different
% portions of the dataset with progressively larger ontologies in terms of \textit{slot-types}. We find that notwithstanding the larger ontologies and the noisiness of our dataset, the NLG performance does not degrade significantly and 
% even improves for some metrics in terms of human evaluation. We also
% observe the importance of using the real world context to improve performances.
% Second, we ask ourselves whether it would be possible to transfer knowledge 
% across QA and task-oriented dialog NLG. To this end we experiment with 
% learning NLG models in a multi-task setting between QA and the popular
% San Francisco restaurant dataset~\cite{wen2015semantically}. Our experiments
% show that learning models in multi-task leads to better performances in terms
% of naturalness for both QA and task-oriented dialog. 

\section{Related work}\label{sec:related}
While classical approaches to NLG involve a pipeline of modules such as content
selection, planning, and surface realization \cite{gatt2018survey}, recently a
large part of the literature investigated end-to-end neural approaches to
NLG. The tasks tackled include dialog, text, and QA. While these tasks share
some similarities, each comes with its own set of challenges and requires
specific solutions.
% VP: Recently a large part of the literature investigated Neural approaches for NLG tasks. The tasks tackled include Dialog, Text and QA. While these tasks share some similarities, each comes with its own set of challenges and requires specific solutions.
% AC: followed VP

%%% First paragraph: COMMON PRACTICES OF NLG FOR DIALOG: Enc-Dec, delexicalization ***
%Current approaches to NLG for dialog:
%%(1) While first approaches to NLG had a pipeline of different tasks (content selection, planning, surface realization etc. \cite{gatt2018survey}), SoA NLG approaches \cite{duvsek2016sequence, nayak2017plan,juraska2018deep} mostly use end-to-end neural Encoder Decoder approach with attention \cite{Bahdanau2014attention}, usually with reranking, as shown by the results of the recent E2E challenge \cite{duvsek2018findings}. The first neural approach to NLG was proposed by ~\citet{wen2015semantically}. (The recent winner of E2E challenge used an ensemble of Encoder Decoders \cite{juraska2018deep}.)
%%(2) Another characteristic of SoA NLG is the use of delexicalization \cite{henderson2014robust}, that is the process of substituting slot values with slot types in the generated text. Delexicalization is used to improve generalization, however recent work shows also the disadavantages of delexicalization and how in some cases lexicalized input is better \cite{nayak2017plan, juraska2018deep}.
%%(3) For a more detailed survey on current NLG approaches, see \citet{gatt2018survey}.
% Maybe (4): Evaluation in NLG: objective metrics (BLEU, SER) and human judgment
\paragraph{NLG for dialog} State of the art NLG models for dialog
\cite{duvsek2016sequence, juraska2018deep} mostly use end-to-end neural
Encoder-Decoder approaches with attention \cite{Bahdanau2014attention} and
re-ranking~\cite{duvsek2018findings}. Ensembling is another technique employed
to boost model performance \cite{juraska2018deep}.  Using delexicalization
\cite{henderson2014robust}, i.e., the process of substituting slot values with
slot types in the generated text, has also shown improvements in many settings.
However, recent work also depicted the disadvantages of
delexicalization~\cite{nayak2017plan}.  In our work, we compare and combine both
delixecalized and lexicalized inputs for the NLG system.

NLG for dialog has been mostly tested in controlled environments using
task-oriented, single domain datasets with limited ontologies \cite{wen2015semantically,novikova2017e2e,balakrishnan2019constrained}.
Although~\citet{wen2016multi} perform multi-domain task-oriented NLG experiments, the
ontologies used are still limited for such settings. Finally, while research has
shown how encoding the previous utterance leads to better
performances~\cite{duvsek2016context}, most settings consider the turns in
isolation \cite{wen2015semantically,novikova2017e2e}.
% VP: the \newcite command look strange here, it should be \cite as before for consistency
% AC: fixed
% VP: Although \newcite{wen2016multi} performed multi-domain task-oriented NLG the ontology used is still limited for such settings. Finally, while research has shown encoding previous utterance leads to better performances \cite{duvsek2016context}, most settings consider the turns in isolation. 
% AC: followed VP

In our work, we perform open-domain NLG with significantly larger ontologies and
also evaluate the impact of adding the context to the input.

\paragraph{NLG for text and QA} 
%%Another related line of research generates text (not dialog) from structured data. Also here some of the main datasets used are set in controlled environments such as weather forecast (WEATHERGOV \cite{liang2009learning}) or robot sportscasting (ROBOCUP \cite{chen2008learning}). Current approaches also use Encoder Decoder with attention \citep{mei2016talk}. Recently more interesting datasets in this area that go beyond controlled environments, such as Wikipedia first sentence generation from Wiki infobox \citet{lebret2016neural}. 
Recent work around NLG for text involves generating text using structured data
using the encoder-decoder networks~\citep{mei2016talk}.  Similarly to dialog,
NLG for text has also been addressed in controlled environments such as weather
forecast \cite{liang2009learning} with few exceptions \cite{lebret2016neural}.
In the literature for QA, most approaches retrieve answers directly or generate
answers jointly with the retrieval, and answers are usually entities or lists of
entities \cite{dodge2015evaluating}. On the contrary, in NLG we assume the
answer has already been retrieved, and the goal is to generate text matching
it. The field of QA which most strictly relates to our work is answer
generation, where current approaches are also based on encoder-decoder networks
encoding information directly from a knowledge base
\cite{yin2016neural,he2017learning,wei2019natural}.
% VP: In the literature for QA, most approaches retrieve answers directly or generate answers jointly with the retrieval. In QA the goal is to obtain a correct answer (i.e. entities or list of entities as answers e.g. WikiMovies \citet{dodge2015evaluating}). On the opposite, in NLG we assume the answer has already been retrieved and the goal is to generate text matching it. The part of QA which most strictly relates to our work is answer generation \cite{yin2016neural,he2017learning,fu2018natural}.
% AC: followed VP
%Some research in this area directly encode triples from the KB, that is simply obtaining entities or list of entities.
% VP: which research? this needs a citation.  What is a KB (knowledge base) it was never introduced. Also the second part of this sentence is non-grammatical.
An additional challenge to answer generation is that there are no publicly
available datasets for this task \cite{fu2018natural}.  

Our approach differs from answer generation in that we structure the task as in NLG dialog literature
with a MR-to-text approach.
% VP: which approaches?
% AC: fixed
%% In our work we pursue a different approach than traditional QA as we attempt to structure the answer retrieval (answer generation?) task as a NLG problem using an MR-to-text approach.
% To the best of our knowledge we are the first to apply the dialog MR-to-text NLG approach to open-domain QA.
% DILEK: Could template-based be better for QA? There is a wide literature on template-based, and it may be useful to mention disadvantages of that approach here. For example the infeasibility of creatign templates with large numbers of slots and their combinations.
% DILEK: In the QA dataset, do we have example sentences that include more than one slot? If so, then template-based won't scale could be the discussion point.
% AC: I looked up a couple of references for template-based, which I was thinking to add to Related work, NLG for dialog section. My plan is to try to fit in first the qualitative error analysis, and then see if we have more space. Anyway we already hint at the non-scalability of template based in the Introduction, so we might also simply add references there if we run out of space.

\section{Datasets}
\label{sec:dataset}
\begin{table}
\scalebox{0.78}{
 \begin{tabular}{c c c c c c c } 
\hline
 & \textbf{Size} & \textbf{Slots} & \textbf{DAs} & \textbf{Words} & \textbf{Domain} & \textbf{Context} \\
\hline\hline
E2E & 51k & 8 & 1 & 2453 & restaurant & no \\
SFX & 5k & 12 & 8 & 438 & restaurant & no \\
\hline\hline
QA.1 & 6k & 147 & 1 & 702 & open & yes \\
QA.2 & 16k & 210 & 1 & 1528 & open & yes \\
QA.3 & 67k & 369 & 1 & 2963 & open & yes \\
\hline
\end{tabular}
}
\caption{Our QA NLG datasets compared to popular (task-oriented) NLG datasets: San Francisco restaurant (SFX) and the NLG E2E challenge (E2E). We report the full size of datasets in terms of MR-text pairs, the number of slot types, DAs, words (computed after delexicalization), domain and whether the dataset comprises the previous utterance or not. % The number of words for each dataset is computed after delexicalization. For E2E the delexicalized slots are \textit{name} and \textit{near} using TGEN.
}
\label{tab:datasets}
\end{table}
%\end{center}
%\vspace{-8.5mm}
% In this section we present the datasets used in our study.
% % VP: this doesn't look like a study, maybe a work?
% In particular, we detail the process of generation of our open-domain QA NLG datasets from a source dataset of QA pairs. Statistics about our QA NLG datasets compared to two task-oriented popular NLG datasets is reported in Table \ref{tab:datasets}.
% VP: add here that the dataset is divided in 3 parts or it is unclear when reading table 2
% AC: cutted this Intro to save space
%
%
%% Chandra Comments: 
% Do we need a footnote and state that we cannot publicly release it? My recommendation is to leave it open say with a real-world conversational agent (and not state/clarify anything). If the reviewer ask about the agent, we can state we didn't add because of confidentiality. Once accepted, we can write the name. 
% \begin{figure*}[h!]\centering
%   %\includegraphics[width=\linewidth]{all_architectures.png}
%   \includegraphics[scale = 0.3, valign=t]{all_architectures.png}\textbf{}
%   \caption{\small Multi-Encoder (DA, Slot type MR, Slot value MR and Context Utterance) with Multi-Attention as the Input and Multi-Decoders (Multiple-Tasks) as the Output}. 
%   \label{fig:all_archictectures}
% \end{figure*}
\begin{figure*}[ht]\centering
  \includegraphics[scale = 0.11, valign=t]{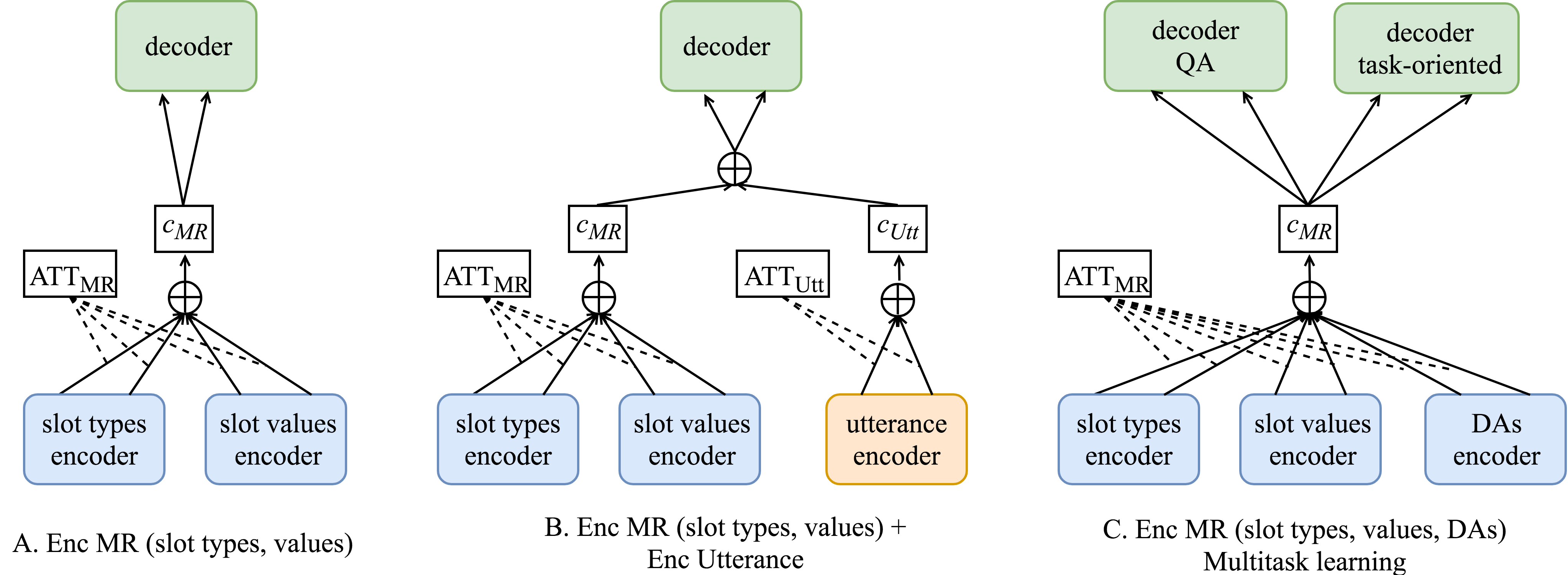}\textbf{}
  \caption{Our baseline model (A) and the models with the previous utterance (B) and for multi-task learning (C). While our baseline model Enc MR (slot types, values) is composed by two encoders for the MR, one for slot types and one for slot values; our model in subfigure B extends this baseline by adding an encoder for the previous utterance. In the multitask learning setting, on the hand, where we do not have the previous context but might have different Dialog Acts (DAs), we add a corresponding encoder (see subfigure C).}
  \label{fig:all_archictectures}
\end{figure*}
\subsection{Question Answering}
\paragraph{Source data}
Our source for generating the MR-text pairs are thousands of open-domain factual
question-answer pairs from commercial data.
The domains covered in this data are manifold, including geography (e.g. `is canada bigger than united states' in Table~\ref{tab:examples}), history (e.g. `when was kentucky founded' in Table~\ref{tab:MRExample}), present-day knowledge (e.g. `will ferrell's wife' in Table~\ref{tab:examples}), grammar (`is there a plural form of pegasus') and even mathematics (`what is one modulo seven'). 
Pairs are grouped according to the type of
question asked. 
Each group consists of a list of specific questions (e.g. ``who
is the wife of barack obama'', ``tell me the wives of henry the viii'') of the
same type (e.g. ``who is the wife of'') asked by real users to a conversational
agent.  Each specific question additionally has: (1) the answer to the question
(e.g. ``michelle obama is obama's wife'') generated by the NLG of the
conversational system, either using information retrieval or a knowledge base
search coupled with templates; (2) relevant noun and verb phrases
(e.g. ``michelle obama'', ``barack obama'', ``wife'') used by the system to
generate the answer, including the ones from the question. Noun phrases are
tagged according to their semantic type (examples of semantic types are
\textit{timepoint} and \textit{human being}), while verb phrases are tagged as
``relation'' types (see ``founded'' tagged as \textit{relStr} in Table
\ref{tab:MRExample}).
% VP: Our question answering dataset is composed by MR-text pairs. This dataset is generated starting from thousands of open-domain factual question-answer pairs asked by real users to a conversational agent. The questions are grouped according to the type of question (e.g. ``who is the wife of barack obama'', and ``tell me the wives of henry the viii'' both belong to the ``who is the wife of'' type). Each question is paired with the answer generated by the NLG of the conversational system (e.g. ``michelle obama is obama's wife''). The conversational agent generates this answers either using information retrieval or querying a knowledge base using templates. The answer-question pairs are further annotated with the relevant noun phrases and verbs utilized by the system to generate the answer, including the ones from the question (e.g., ``michelle obama'', ``barack obama'', ``wife''). The noun phrases are tagged according to their semantic type (e.g., \textit{timepoint} and \textit{human being}), while verbs are tagged as ``relation'' types (for example ``founded'' is tagged as ``rel'' in Table \ref{tab:MRExample}).
% VP: at this point it is still unclear how you go from natural language to MR 

The answers in the source data are varied, and range from a simple entity to a
fully formed answer, as in Table~\ref{tab:MRExample} example where valid answers
to the question ``when was kentucky founded'' can be ``1792'' or ``kentucky
formed in 1792''.  This shows an interesting difference between our QA data and
task-oriented NLG datasets. While for task-oriented NLG all valid responses for
a single MR have the same slot types (i.e., the ones in the input MR), in our
dataset this is not always true.
% VP: The answers in the source data are varied, and range from a simple entity to a fully formed answer. Table~\ref{tab:MRExample} shows two examples of answers to the question ``when was kentucky founded''. The answer can be simply ``1792'' or ``kentucky formed in 1792''. This example shows an interesting difference between our QA data and task-oriented NLG datasets. While for task-oriented NLG datasets all valid responses for a single MR have the same slot types (i.e., the ones in the input MR), in our dataset this is not always the case.
% AC: followed VP

\paragraph{QA NLG datasets} We generate the NLG input-output pairs for QA from our
source data. In order to perform cross-application experiments, we maintain the
same MR-text format as task-oriented dialog NLG.  The target output is the text
of the answers in the source data. To generate the input MRs we assumed only one
DA across all answers, i.e. ``inform''; for the slots, we used the semantic
types and relations for noun phrases and verb phrases in the source data as slot
types, while the actual entity or verb was used as the corresponding slot
value. \footnote{Although we use the original tags of the source data, a similar
  representation could be produced by tagging noun phrases with their Named
  Entity type and verb phrases with a ``relation'' slot type.}  On top of the
generated MR we use, as additional input, the previously asked question as
context.

Answers are delexicalized~\cite{henderson2014robust} to improve generalization.
Since we do not have alignment between entities in the input and the generated
text, we use a heuristic-based aligner which we also use to filter out data that
could not be appropriately aligned. All noun phrases are delexicalized while
verb phrases are not.  Furthermore, similar to \cite{juraska2018deep}, we use
delexicalization for data augmentation.  We generate additional references for
each MR, besides the original one, by considering all delexicalized answers in
the question group as candidate template answers for each specific question in
the group and then substituting (where possible) slots values which are already
available in the input.  The text of the previous question is also
delexicalized.

Finally, to investigate performances across different ontology sizes, we
generate 3 different partitions of the data (QA.1, 2 and 3 in Table
\ref{tab:datasets}) with a progressively larger number of slot types.  Each QA
partition was split in train, test and development set (using a 80-10-10 split)
according to the type of question asked. We ensured there was no overlap between
the different sets to test if models generalize to previously unseen questions.
% TODO: report number of delex "templates" for each partition (averages QA.1: 1.9, QA.2: 3.2, QA.2:5 )

% VP is the 'and' needed?
% AC: fixed

% \subsection{E2E challenge dataset}
% The E2E NLG dataset \cite{novikova2017e2e} is an NLG dataset in the restaurant booking domain.
% \textbf{Data preprocessing} delexicalized everything except binary values

\subsection{Task-oriented Dialog}
As a task-oriented NLG corpus for our multi-task learning experiments we use the
popular San Francisco restaurants (SFX;~\citet{wen2015semantically}) dataset. %, since it is one of the most popular datasets used for benchmarking NLG models.
Statistics about the dataset is shown in Table \ref{tab:datasets}.  Although SFX
is not large (6k examples), compared to the E2E NLG corpus it presents more
% rahul-inlg: We use-> For all our datasets, we use
variation for DA (although less in style). For all our datasets, we use the TGEN
library \footnote{\url{https://github.com/UFAL-DSG/tgen}}
\citep{duvsek2016sequence} to delexicalize all slot types except binary values.

%, another reason why we chose it for our experiments.
% VP: you delexicalize SFX, E2E, or both? not clear to me but I'm not familiar with the datasets
%
% \paragraph{Data preprocessing} All slot types were delexicalized, except for binary values, using 
% \begin{figure*}[h!]\centering
%   %\includegraphics[width=\linewidth]{all_architectures.png}
%   \includegraphics[scale = 0.3, valign=t]{all_architectures.png}\textbf{}
%   \caption{\small Multi-Encoder (DA, Slot type MR, Slot value MR and Context Utterance) with Multi-Attention as the Input and Multi-Decoders (Multiple-Tasks) as the Output}. 
%   \label{fig:all_archictectures}
% \end{figure*}
% \section{Model and Architectures} \label{sec:seq2seq}
% We tried variety of architectures and inputs for training the NLG models for multiple settings. The following sections depict the model architectures which we tried.

\section{Model and Architectures} \label{sec:seq2seq}
In this section we present the variety of different architectures used in our experiments. Although all our models are based on the Encoder-Decoder framework, we investigate architectures with different number of Encoders (up to 3). Given this variety, for clarity, we follow a template Enc $<$Encoder type$>$ for naming our different models. The type of Encoder, in particular, can be of Meaning Representation (MR) type, when we encode parts of the MR, such as slot types, values or Dialog Act; or it can be of Utterance type, when we encode the previous utterance context. 

\paragraph{Encoder-Decoder with Attention}  %\citet{juraska2018deep} recently achieved state-of-the-art results on variety of task-oriented datasets using an encoder-decoder model with attention. %~\cite{Bahdanau2014attention}.
Following recent state-of-the-art approaches to NLG for dialog
\cite{juraska2018deep,balakrishnan2019constrained}, our models are based on the
Encoder-Decoder with Attention framework.  In particular, we use bidirectional
Gated Recurrent Units (GRU) and Luong general
attention~\cite{luong2015effective} as our baseline. 
While we also experimented  with other types of architectures, such as using Long-Short-Term Memory Units~\cite{hochreiter1997long} instead of GRUs and different types of attention (including Bahdanau attention~\cite{Bahdanau2014attention} and Luong dot attention~\cite{luong2015effective}, this combination gave us the best results for our setting. 
Depending on the encoder
used, either slot type or slot value, we refer to this model as Enc MR (slot
types) or Enc MR (slot values).

\paragraph{Multi-Encoder, Single Decoder:} We expand the baseline (Enc MR) models
using multiple inputs from the MR (slot types, values, DAs), each encoded by a
different encoder.  The attention is performed on their concatenated output to
produce the MR context vector $c_{MR}$.  Figure \ref{fig:all_archictectures} A
shows an example of such an architecture using two encoders, one for slot types
and one for slot values.  Furthermore, we experimented with adding the previous
utterance as input with an additional encoder (Enc Utterance).  In this case,
the context vector for the previous utterance $c_{Utt}$ is produced by an independent attention
mechanism and the outputs of both attentions ($c_{MR}$ and $c_{Utt}$) are
concatenated (see Figure \ref{fig:all_archictectures} B).

%\subsection{Multi-Encoders and Multi-Decoder (ME-MD)}\label{ssec:multi_enc_multi_dec}
\paragraph{Multi-Encoder, Multi-Decoder:} We also performed multi-task learning,
jointly training the models for both QA and task-oriented NLG.  As shown in
Figure~\ref{fig:all_archictectures} C, we shared the encoders and corresponding
input layers across multiple tasks while we maintained multiple decoders for
% rahul-inlg -> added the last sentence to explain multi-tasking a bit, can be 
% explanied better
individual tasks. We alternated between mini-batches from various data sources to
perform multitasking. 
% VP: Figure~\ref{fig:all_archictectures} shows the topology of this model with a shared encoders and input layers across multiple domains and multiple decoders for each individual task.
% VP: The wording is confusing as we talk about multiple shared encoders. If I understood correctly there is a single encoder for each input (e.g., slot types, slot values,...) but the encoder is shared by multiple tasks. In the text (also the one I suggested) this is not very clear.
% In our multi-task setting, we do not u previous utterance therefore, we shared a single Attention layer across the DAs (since there is not a one-to-one mapping of DAs across multiple domains), slot type and slot value MRs.
% VP: is Slot-value Meaning Representation all one thing? or you need a comma?

%\begin{figure}
%%\begin{figure*}[h]\centering
% \includegraphics[width=\linewidth]{multi_enc_multi_dec_2.png}
  %%\includegraphics[scale = 0.7, valign=t]{multi_enc_multi_dec.png}
%  \caption{\small Multi-Encoder (Slot type MR, Slot value MR and Context Utterance) with %Multi-Attention as the Input}
%  \label{fig:multi_enc_multi_dec}
%\end{figure}
%%\end{figure*}

%General encoder decoder architecture
%Dot vs general Luong attention \cite{luong2015effective}
%Teacher forcing
%Reranker?

%\subsection{Multiencoders}

%\subsection{Multidecoders}
%Table with architecture figures
% TODO: collective figure with 3 architectures :
% 1: Enc (slot types) + Dec (words)
% 2: Enc (slot types, values, das) + Enc (context) + Dec (words)
% 3: Enc (slot types, values, das) + Enc (context) + MultiDec (words)

\section{Evaluation}\label{sec:eval}

As word overlap metrics may not have a good correlation with human judgment for
NLG output evaluation \cite{stent2005evaluating}, we use both objective metrics
and human evaluation. 
\paragraph{Objective metrics} Besides the standard BLEU
score (obtained using the official E2E NLG challenge evaluation script
\footnote{We do not report other word overlap metrics (e.g., METEOR) computed by
  the E2E evaluation scripts due to space limitations and correlation with the
  BLEU score.}), we report different types of Slot Error Rate (SER). In dialog
NLG approaches SER shows the number of correct slots in the output compared to
the input MR. We refer to this metric as SER\textsubscript{mr} to differentiate
it from its modified versions we introduce next.  The formula
\cite{wen2015semantically} is:
\begin{equation}
SER_{mr} = \frac{p_{mr}+q_{mr}}{N_{mr}}
\end{equation}
where $N_{mr}$ is the total number of slots in the input MR and $p_{mr}$ ,
$q_{mr}$ are respectively the number of missing and redundant slots in the
output.  This formula works well for task-oriented NLG approaches, but it
assumes a one-to-one relationship between the slots in the input MR and the
output text.  We found this assumption might not hold for our QA datasets where
not all slots in the input MR need to be realized for the output to be
correct. An example of this is shown in Table \ref{tab:MRExample}, where the
first QA reference text (`1792') would be penalized with 3 missing slots, while
still being correct.

% VP: This formula works well for task based NLG approaches but it assumes a one-to-one relation between the slots in the input MR and the one in the output pairs. We found this assumption might not hold for our QA datasets where not all the slots from the input MR need to be realized for the output to be correct. As an example consider QA MR shown in Table \ref{tab:MRExample}, the first text output would be penalized with 3 missing slots. 
% AC: followed VPs

In order to capture this different behaviour we designed additional NLG metrics
tailored for QA. Slot Error Rate Target (SER\textsubscript{trg}) is a
modification of SER\textsubscript{mr} where we simply substitute the MR with the
main reference text:
\begin{equation}
SER_{trg} = \frac{p_{trg}+q_{trg}}{N_{trg}}
\end{equation}
SER\textsubscript{trg} is designed to penalize both missing and redundant slots
compared to the target sentence. Hence, using SER\textsubscript{trg} the first
QA reference text in Table \ref{tab:MRExample} would not be penalized.

Slot Error Rate MultiTarget (SER\textsubscript{mtrg}), on the other hand,
penalizes redundant slots that did not appear in any of the references:
\begin{equation}
SER_{mtrg} = \frac{p_{mtrg}}{N_{mtrg}}
\end{equation}
where $N_{mtrg}$ are all slots appearing in any reference and $p_{mtrg}$ are the
slots in the output that did not appear in any reference sentence.
% VP: where $N_{mtrg}$ is the number of slots appearing in any reference and $p_{mtrg}$ is the number of slots in the output that did not appear in any reference sentence.
% VP: is my understanding of $p_{mtrg}$ correct?
% AC: yes
%So, in the case where our model's output was ``kentucky formed in year 1792'' given the QA MR in Table \ref{tab:MRExample} and we had as main reference the first QA reference (``1792'') but the second QA reference in the pool of possible text references (``kentucky formed in 1792''), the output would still be considered correct. \\
To compute SER\textsubscript{mtrg} for the model output ``kentucky formed in
1792'' given the QA MR in Table \ref{tab:MRExample} we assume to have two
references ``1792'' and ``kentucky formed in 1792''. In this case,
SER\textsubscript{mtrg} would consider the output correct as all of its slots
appear in at least one of the references.

\begin{table*}[htb]
%\centering
\setlength{\tabcolsep}{2.8pt}
% \small
\scalebox{0.78}{
\begin{tabularx}{\textwidth}{l cccc|cccc|cccc}
\cline{1-13}
 & \multicolumn{4}{c|}{\textbf{QA.1}}  & \multicolumn{4}{c|}{\textbf{QA.2}} & \multicolumn{4}{c}{\textbf{QA.3}} \\ 
 %\cline{2-13} 
 & \multicolumn{1}{l}{\textbf{BLEU}} & \multicolumn{1}{l}{\textbf{\begin{tabular}[c]{@{}l@{}}SER\textsubscript{mr}\end{tabular}}} & \multicolumn{1}{l}{\textbf{\begin{tabular}[c]{@{}l@{}}SER\textsubscript{trg}\end{tabular}}} & \multicolumn{1}{l|}{\textbf{\begin{tabular}[c]{@{}l@{}}SER\textsubscript{mtrg}\end{tabular}}} & \multicolumn{1}{l}{\textbf{BLEU}} & \multicolumn{1}{l}{\textbf{\begin{tabular}[c]{@{}l@{}}SER\textsubscript{mr}\end{tabular}}} & \multicolumn{1}{l}{\textbf{\begin{tabular}[c]{@{}l@{}}SER\textsubscript{trg}\end{tabular}}} & \multicolumn{1}{l|}{\textbf{\begin{tabular}[c]{@{}l@{}}SER\textsubscript{mtrg}\end{tabular}}} & \multicolumn{1}{l}{\textbf{BLEU}} & \multicolumn{1}{l}{\textbf{\begin{tabular}[c]{@{}l@{}}SER\textsubscript{mr}\end{tabular}}} & \multicolumn{1}{l}{\textbf{\begin{tabular}[c]{@{}l@{}}SER\textsubscript{trg}\end{tabular}}} & \multicolumn{1}{l}{\textbf{\begin{tabular}[c]{@{}l@{}}SER\textsubscript{mtrg}\end{tabular}}} \\ 
\cline{1-13} 
% rahul-inlg -> change 2-enc to 1
Enc MR (slot types, values) & 0.85  & \textbf{0.42}  & 0.21 & 0.023  & 0.77 & 0.44 & 0.3 & 0.057 & 0.66 & \textbf{0.43} & 0.37 & \textbf{0.05} \\
+ Enc Utterance delex  & 0.89 & 0.44 & 0.19 & 0.014 & 0.83 & \textbf{0.43} & 0.23  & \textbf{0.025} & 0.7 & 0.45 & 0.32 & 0.054 \\
+ Enc Utterance lex  & \textbf{0.95} & 0.46 & \textbf{0.15} & \textbf{0.012} & \textbf{0.89} & 0.47 & \textbf{0.19} & 0.03 & \textbf{0.72} & 0.46 & \textbf{0.28} & 0.067   \\
\cline{1-13}
\end{tabularx}
}
% DILEK: It would be useful to include significance ranges, especially if the tools you use for eval already support it.
% AC: unfortunately I don't think we'll manage to do that. For the objective evaluation, as far as I recall the evaluation script from E2E challenge didn't have significance testing. For human evaluation, we did significance tests, however tests were significant only for QA.1 (where human evaluation results are farther apart).
\caption{Objective metrics results on three QA NLG datasets with increasingly larger ontologies. The models under comparison are a baseline with two encoders, for slot types and slot values, and its extensions with a delexicalised or lexicalised previous utterance. While for BLEU score the higher the better, for all types of Slot Error Rate (SER) the lower the better.}
\label{tab:QAObjResults}
\end{table*}
\begin{table*}[htb]
\centering
\setlength{\tabcolsep}{8pt}
\scalebox{0.78}{
\begin{tabular}{l cccc|cccc|cccc}
\cline{1-13}
%\begin{tabularx}{\columnwidth}{l|cclc|cclc|cclc}
 & \multicolumn{4}{c|}{\textbf{QA.1}}  & \multicolumn{4}{c|}{\textbf{QA.2}}   & \multicolumn{4}{c}{\textbf{QA.3}}  \\ 
 %\cline{2-13} 
%\multicolumn{1}{l|}{}                             & \textit{\textbf{Nat.}} & 
                         & \textit{\textbf{Nat.}} &
\textit{\textbf{Inf.}} & \textit{\textbf{Conv.}} & \textit{\textbf{Ans.}} & \textit{\textbf{Nat.}} & \textit{\textbf{Inf.}} & \textit{\textbf{Conv.}} & \textit{\textbf{Ans.}} & \textit{\textbf{Nat.}} & \textit{\textbf{Inf.}} & \textit{\textbf{Conv.}} & \multicolumn{1}{c}{\textit{\textbf{Ans.}}} \\ \cline{1-13} 
%\multicolumn{1}{l|}{2-ENC MR(slot types, values)} & 3.73                   & 3.8            
% rahul-inlg -> I changed 2-enc to 1-enc
Enc MR (slot types, values) & 3.73                   & 3.8 
& 3.96                    & 0.36                         & 3.88                   & 3.78                   & 4.57                    & 0.38                         & 4.29                   & 4.37                   & 4.40                    & 0.73   \\
%\multicolumn{1}{l|}{+ 1-ENC UTT delex}            & 4.61                   & 4.63           
+ Enc Utterance delex            & 4.61                   & 4.63
& \textbf{4.59}           & 0.67                         & 4.64                   & 4.32                   & \textbf{4.88}           & 0.5                          & 4.52                   & 4.47                   & \textbf{4.51}           & 0.79                         \\
%\multicolumn{1}{l|}{+ 1-ENC UTT lex}              & \textbf{4.7}           & \textbf{4.69}  
+ Enc Utterance lex              & \textbf{4.7}           & \textbf{4.69}
& 4.48                    & \textbf{0.78}                & \textbf{5.15}          & \textbf{4.88}          & 4.85                    & \textbf{0.67}                & \textbf{4.57}          & \textbf{4.57} & 4.45 & \textbf{0.80} \\
\cline{1-13}

\end{tabular}
}
\caption{Human evaluation results on three QA NLG datasets with increasingly larger ontologies. The models reported are a baseline with two encoders, for slot types and slot values, and its extensions with a delexicalised or lexicalised previous utterance. 
We report averages of Naturalness (\textit{Nat.}), Informativeness (\textit{Inf.}), and how conversational the response was judged (\textit{Conv.}) on a scale of 1 to 6. Additionally, we report the average of whether responses could be considered an answer to the given question (\textit{Ans.}), given to annotators as a binary choice.}
\label{tab:QASubjResults}
\end{table*}

\begin{table}[htb]
%\small
%\centering
\setlength{\tabcolsep}{4pt}
\scalebox{0.88}{
%\begin{tabular}{l|c|ccc}
\begin{tabularx}{\columnwidth}{l c cccc}
\cline{1-6}
& \multicolumn{1}{l}{\textbf{Dataset}} & \textbf{BLEU}     & \textbf{SER\textsubscript{mr}} & \textbf{SER\textsubscript{trg}} & \textbf{SER\textsubscript{mtrg}} \\ 
\cline{1-6} 
baseline & \multirow{2}{*}{SFX} & 0.727 & \textbf{0.40} & - & - \\
%\multicolumn{1}{l|}{+ QA.3}   &                                       & \textbf{}       
+ QA.3 &  & \textbf{0.74} & 0.413 & - & - \\ 
\cline{1-6} 
%\multicolumn{1}{l|}
{baseline} & \multirow{2}{*}{QA.3}  & 0.659 & \textbf{0.429} & 0.37 & \textbf{0.05} \\
+ SFX  & & \multicolumn{1}{l}{\textbf{0.673}} & 0.44 & \textbf{0.368} & 0.07 \\
\cline{1-6}
\end{tabularx}
%\end{tabular}
}
\caption{Objective metrics of multitask learning experiments combining QA (QA.3) and task-oriented dialog (SFX) NLG. For all Slot Error Rate (SER) metrics the lower the better.}
\label{tab:multiObjResults}
\end{table}
%\end{center}
% \vspace{-1.5mm}
\begin{table}[htb]
%\small
\setlength{\tabcolsep}{8pt}
%\center
\scalebox{0.88}{
%\begin{tabular}{|l|c|cccc|}
\begin{tabularx}{\columnwidth}{l c cccc}
 \cline{1-6}
         & \multicolumn{1}{l}{\textbf{Dataset}} & \textit{\textbf{Nat.}}            & \textit{\textbf{Inf.}} & \textit{\textbf{Conv.}} & \multicolumn{1}{l}{\textit{\textbf{Ans.}}} \\ \cline{1-6} 
%\hline
baseline & \multirow{2}{*}{SFX}                  & 4.69                              & \textbf{5.50}          & -                       & -                                                \\
+ QA.3   &                                       & \textbf{5.11}                     & 5.40                   & -                       & -                                                \\ \cline{1-6}  %\hline
baseline & \multirow{2}{*}{QA.3}                 & 4.29                              & 4.33                   & 4.40                    & \textbf{0.73}                                    \\
+SFX     &                                       & \multicolumn{1}{l}{\textbf{4.43}} & \textbf{4.38}          & \textbf{4.5}            & 0.72  \\
\cline{1-6}
%\end{tabular}
\end{tabularx}
}
\small \caption{Results of multitask learning experiments combining NLG for QA (on QA.3) and task-oriented dialog (on SFX) according to human evaluation.}
\label{tab:multiHumanResults}
\end{table}

\paragraph{Human evaluation} In all experiments, for each dataset, we selected a
sample of 100 MR-text pairs from the test set.  Pairs were randomly selected
among those where all models under comparison in the experiment had generated
different output text.  Data for all reported experiments were annotated by 2
human annotators, and final ratings were averaged between the two.  In all
experiments annotators, presented with MR and all outputs of the systems under
comparison, were asked to rate the \textit{naturalness} and
\textit{informativeness} of the generated output using a 1-6 Likert score, as in
previous NLG dialog evaluations \cite{gatt2018survey}.  Additionally, for the QA
datasets annotators had also the previous question as context.  Moreover, for
the QA datasets annotators were asked to rate how \textit{conversational} the
output was, on the same Likert scale, and whether or not the output could
ultimately be considered an answer to the question (\textit{answer}), as a
binary choice. %These two metrics were added to capture other factors which might have an influence on the quality of the generated response.
\section{Experimental setup}
The hyperparameters chosen for our models were empirically determined through various experiments.
Both encoder and decoder in all our models had only one layer, as we noticed additional layers did not give improvements. 
All embeddings were trained from scratch with a fixed dimension of 50.
Models were trained using a cross-entropy loss function and the Adam~\cite{kingma2014adam} optimizer with a learning rate of 0.001, for 1000 epochs, with early stopping on the validation set. We used mini-batches of size 32.

For the NLG models for QA, experiments on QA.1 (not reported due to space limitations) 
with different encoders combinations showed that the best performances were achieved using all input types (slot type, value, and previous context) with lexicalized ($+$ Enc Utterance lex) or delexicalized ($+$ Enc Utterance delex) previous context in terms of all metrics, except SER\textsubscript{trg}. On this metric, the architecture with slot types and values, but without the previous context (Enc MR (slot types, values)) achieved the best performance (cf. Table \ref{tab:QAObjResults}). For this reason, we chose to report the performances of these architectures in our QA experiments.

\section{Results}

\paragraph{Open domain QA}

In our first batch of experiments we test various Encoder-Decoder architectures
on our 3 different partitions of QA NLG data.

As we can see from Table \ref{tab:QAObjResults}, in general, the best
performances across all QA datasets for both BLEU and SER\textsubscript{trg} are
achieved by the model using as additional input the lexicalized previous
question, followed by the model with the delexicalized one.  However,
SER\textsubscript{mr} results show the opposite picture, where the baseline with
only slot types and values performs better (except for QA.2 where the score is
close to the model with the delexicalized input) and the model with the
lexicalized previous utterance is the worst.  SER\textsubscript{mtrg} shows, on
the other hand, that the context might slightly degrade performances with bigger
ontologies in terms of all text references.

\begin{table*}[htb]
\centering
\scalebox{0.65}{
% \begin{tabularx}{\textwidth}{l cccc|cccc|cccc}
 \begin{tabular}{c|cc|cccc} 
\hline
\textbf{Dataset} & \multicolumn{2}{c|}{\textbf{Input}} & \textbf{baseline} & \textbf{$+$delex} & \textbf{$+$lex} & \textbf{multitask} \\
& \textbf{context} & \textbf{MR} & & & & \\
\hline\hline
QA.1 & `sing your song & \textit{human being}: & `sing your song & `sing your song & `vic ruggiero .' & - \\
 & writer' & 'vic ruggiero',  & ' coach is & ' writer is & & \\
  & & \textit{a}:'sing your song' & vic ruggiero .' & vic ruggiero .' & & \\
  & & \textit{b}:'writer' &  &  & & \\
\hline
QA.1 & `did abraham & \textit{true}:'positive polarity', & `yes . abraham & `yes , abraham & `yes , abraham & - \\
 & lincoln have & \textit{lp}:'dad' & lincoln has at &  lincoln had & lincoln had & \\
  & a dad' & \textit{ro}:'abraham lincoln' & least one dad .' & a mother.' & a father.' & \\
\hline
QA.1 & `what is the & \textit{timepoint}:'1999', & `1999 's starting & `the the masters & `the masters & - \\
 & masters & \textit{a}:'the masters' & date point & 's starting date & was created & \\
  & starting date' & \textit{b}:'starting date' & is 1999 .' & point is 1999 .' & on 1999 .' & \\
\hline\hline
QA.3 & `is canada & \textit{false}:'negative polarity', & `no , canada is not & `no , canada is not & `no , canada is not & `no , canada is not \\
&  bigger than  & \textit{r}:'bigger than',  & the bigger than & bigger than & bigger than & bigger than \\
& united states'  & \textit{y}:'united states', & united states .' & united states .' & united states .' & united states .' \\
& & \textit{x}:'canada' & & & & \\
\hline
QA.3 & `will ferrell's  & \textit{human being}:  & `will ferrell 's 's wife & `will ferrell 's 's wife & `viveca paulin .' & `will ferrell 's wife\\
& wife' & 'viveca paulin',  & is viveca paulin .' & is viveca paulin .' &  &  is viveca paulin .' \\
& & \textit{a}:"will ferrell 's", & & & & \\
& & \textit{b}:'wife' & & & & \\
\hline
QA.3 & `popsicle & \textit{business}:'unilever', & `popsicle 's maker & `popsicle 's  & `unilever .' & popsicle 's maker \\
&   maker' & \textit{a}:'popsicle' & is unilever .' & manufacturer  &  & is unilever .' \\
& & \textit{b}:'maker' & & is unilever .' & & \\
\hline\hline
SFX & - & \textbf{inform} & sanjalisco allows & - & - & sanjalisco allows \\
 &  & (\textit{name}:`sanjalisco', & kid -s and &  &  & kid -s \\
 &  & \textit{kidsallowed}:`yes') & is located &  &  & \\
\hline
SFX & - & \textbf{inform} & red door cafe is a & - & - & red door cafe is a \\
 &  & (\textit{name}:`red door cafe', & nice restaurant in &  &  & nice restaurant in \\
 &  & \textit{area}:`cathedral hill') & the cathedral hill  &  &  & cathedral hill that is \\
  &  & \textit{goodformeal}:`breakfast') & does not allow kid -s and &  &  &  good for breakfast and \\
  &  & \textit{kidsallowed}:`no') & is good for breakfast &  &  &  does not allow kid -s \\
\hline
SFX & - & \textbf{inform} & darbar restaurant is & - & - & darbar restaurant is \\
 &  & (\textit{name}:`darbar restaurant', &  a pakistani restaurant &  &  & a nice restaurant that \\
 &  & \textit{food}:`pakistani') &  that allows kid -s and &  &  & serves pakistani food \\
  &  & \textit{goodformeal}:`lunch') & is good for lunch &  &  & and allows kid -s  \\
  &  & \textit{kidsallowed}:`yes') &  &  &  &   \\
\hline
\end{tabular}
}
\caption{Examples of different outputs from our models when given the same input Meaning Representation (and previous context when available) on two of our Question-Answering datasets (QA.1, QA.3) and on a task-based (SFX) dataset.}
\label{tab:examples}
\end{table*}

Human evaluation, on the other hand, seems in line with the picture depicted by
BLEU and SER\textsubscript{trg}. Table \ref{tab:QASubjResults} shows the model
with the lexicalized context is regarded as the best, closely followed by the
model with the delexicalized one in every metric except for
\textit{conversational}, where delexicalized is better.  This confirms our
hypothesis that SER\textsubscript{mr} might be a less reliable metric to
evaluate NLG QA output.  Moreover, although we notice a consistent but not
drastic degradation in terms of BLEU and SER\textsubscript{trg} in correlation
with bigger ontologies, human evaluation shows an even more gentle degradation
between QA.1 and 3 for many metrics. Interestingly, it seems the ability of all
models to give a proper answer to the question (\textit{answer}) increases from
QA.1 to 3.

\paragraph{Multitask learning}
In our multitask learning experiments we combine the biggest QA dataset, QA.3,
with a task-oriented corpus, SFX.  We aim to investigate the possibility of
transferring knowledge across different NLG systems, notwithstanding the
diversity of the data in terms of domain, ontology size, DAs, application (QA
vs. task-oriented). Since context is not available in SFX, the model we use has
3 MR Encoders (slot types, values, DAs) and 2 Decoders (one for each task).

Our experiments show that the NLG QA task improves the fluency on SFX both in
terms of objective metrics (in Table \ref{tab:multiObjResults}) and human
evaluation (in Table \ref{tab:multiHumanResults}). However, training with QA
seems to slightly degrade the model efficiency in generating the correct slots.
% VP: This sentence is vague. Maybe quantify the improvement in fluency in terms of % increase?
This is to be expected given the difference in the relation between slots in MR
and output (one-to-one in SFX, variable in QA.3).  As for QA.3 results, it seems
the task-oriented NLG task improves QA NLG performances in terms of fluency
(BLEU and \textit{Naturalness}) and slot errors (SER\textsubscript{trg} and
\textit{Informativeness}). SER\textsubscript{mr} and SER\textsubscript{mtrg},
however, show a slight degradation.
% The baseline model was the one with most grammatical errors for the QA task (e.g. ``will ferrell 's 's wife is viveca paulin'',  or ``no , canada is not the bigger than united states .''). Lexicalizing the previous question improved the grammar and produced short correct answers (e.g. `unilever.'). Delexicalizing the input produced more conversational but also more factually incorrect answer (e.g. `popsicle's manufacturer is unilever').
We observe task-oriented NLG also makes QA NLG more conversational, however
slightly reducing its probability of being an answer to the posed question as
well.% A qualitative analysis confirmed responses of both QA and task-oriented models trained in multitask were more grammatical.

Finally, comparing all experiments on QA.3, we notice that although multi-task
learning helps, the previous context (either lexicalized or delexicalized) plays
a critical role in improving the overall performance.

%\begin{center}
% AC: \begin{table}[H]

\section{Qualitative analysis}

In this section we report the qualitative analysis we performed on the human annotated testset.
Table \ref{tab:examples} reports some output examples from different models
given the same input MR. In particular, we are interested on the impact of adding various features and multi-tasking.

% QA
\paragraph{QA}
According to our qualitative analysis on the QA datasets, the baseline model is the one with most grammatical errors (e.g. ``will ferrell 's 's wife is viveca paulin'', ``no ,
canada is not the bigger than united states .''), while in general adding
``delex'' and ``lex'' features generates more grammatical responses. This observation was confirmed from both the objective (in terms of BLEU score) and subjective (\textit{naturalness}) evaluations performed.

We also notice how lexicalizing the previous question helps in producing generally correct
(e.g. `unilever.') however shorter answers, which can be regarded as less
conversational. 
Delexicalizing the input, on the other hand, produces more conversational
(e.g. `popsicle's manufacturer is unilever') but also more factually incorrect
answers. 
These observations seem also in line with the subjective evaluation results, which on average reported the best scores for the model with lexicalized previous context ($+$ Enc Utterance lex) on whether the generated text could be considered an answer to the given question (\textit{answer}), except for the metric rating how \textit{conversational} the output was, for which the model with delexicalized previous context ($+$ Enc Utterance delex) was regarded as the best one across all QA partitions.
% Additional examples of generated output are available in the supplementary material.
% VP: ??? If there is a degradation in terms of BLEU and SER, and the same degradation (although more gentle) appears in the human evaluation how do we conclude that the ability to answer increases from QA.1 to QA.3? Is this compared to the baseline? Also if comparing to the baseline Table 3 and 4 should probably have a line for the baseline.
% AC: Our baseline is 1-Enc MR (slot types, values), the ability to answer is given by the human evaluation (Answ) in Table 4
% On QA1 and QA3 dataset, adding ``delex'' and ``lex'' features generates more grammatical responses. 
%It can also be observed that ``lex'' features tends to prefer shorter but correct responses, which is less conversational. 

% Multitask
\paragraph{Multitask}
Looking at the output of the models trained in a multi-task learning setting, we observe that the baseline tends to be more prone to grammatical errors compared to models jointly trained with another task (e.g. in Table \ref{tab:examples} `sanjalisco allows kid-s and is located'). Due to multi-tasking the models generate more grammatically correct and natural responses for both SFX and QA.3.

% \section{Discussion}\label{sec:discussion}
% Importance of adapting evaluation metrics.
% Importance of understanding the importance of some entities compared to other ones. (future work)

\section{Conclusions}\label{sec:conclusions}

In this work, we apply the traditional dialog MR-to-text approach to NLG to an
open-domain QA setting, with sensibly larger ontologies compared to current task-oriented dialog approaches.  Our goal was to test the
reliability of current approaches to NLG for dialog in an environment where the number of
slots could be substantial, a requirement that is critical to meet if we want to
move towards an integrated NLG module across different domains.

The experiments presented show the feasibility of learning a NLG module for QA
using a MR-to-text approach. NLG models performances on datasets with
progressively bigger ontologies reported a continuous but not drastic decline
for most metrics.  Moreover, our multitask learning experiments showed that
learning NLG models jointly for QA and task-oriented dialog improves single
tasks performances in terms of fluency.  Results across different experimental
settings also point towards the vital role played by the previous utterance
context (delexicalized and especially lexicalized) to improve NLG models for
open-domain QA. % Although this finding was to be expected, it is still not a common practice in NLG for dialog datasets to include the previous utterance.

While we envision our approach as a first step towards an integrated statistical NLG module for a dialog system, still much remains to be done in order to achieve such a challenge. In this work, for example, we saw the importance of adapting approaches to NLG typical of task-oriented dialog when moving to an open-domain QA setting. This is important not only in terms of modelling (the essential role of the previous utterance), but also in terms of evaluation (designing metrics able to capture the relative importance of some slots in a given answer compared to others).

As our future work, we would like to expand our multi-task learning experiments to novel NLG datasets, for example recently proposed datasets of reviews \cite{oraby2019curate}. Another possibility would be to explore transfer learning, rather than multi-task learning for NLG in the MR-to-text approach.
Additionally, another interesting research direction would be the investigation of evaluation metrics for NLG in a QA setting, for example to better capture the centrality of some slots (or entities) compared to others when answering a given question.
%, and how additional information could improve how conversational a given answer is perceived.

\bibliography{naaclhlt2019}
\bibliographystyle{acl_natbib}

\end{document}